\documentclass{article}
\usepackage{ijcai16}
\usepackage{graphicx}
\usepackage{amsmath}
\usepackage{diagbox}

\usepackage{times}
\title{Model-based Deep Hand Pose Estimation}
\author{Xingyi Zhou\textsuperscript{1}, Qingfu Wan\textsuperscript{1}, Wei Zhang\textsuperscript{1}, Xiangyang Xue\textsuperscript{1}, Yichen Wei\textsuperscript{2} \\ \textsuperscript{1}Shanghai Key Laboratory of Intelligent Information Processing \\School of Computer Science, Fudan University\\ \textsuperscript{2}Microsoft Research \\ \textsuperscript{1}\{zhouxy13, qfwan13, weizh, xyxue\}@fudan.edu.cn, \textsuperscript{2}yichenw@microsoft.com}

\begin{document}

\maketitle

\begin{abstract}
  Previous learning based hand pose estimation methods does not fully exploit the prior information in hand model geometry. Instead, they usually rely a separate model fitting step to generate valid hand poses. Such a post processing is inconvenient and sub-optimal. In this work, we propose a model based deep learning approach that adopts a forward kinematics based layer to ensure the geometric validity of estimated poses. For the first time, we show that embedding such a non-linear generative process in deep learning is feasible for hand pose estimation. Our approach is verified on challenging public datasets and achieves state-of-the-art performance.
\end{abstract}

\section{Introduction}
Human hand pose estimation is important for various applications in human-computer interaction. It has been studied in computer vision for decades~\cite{erol2007vision} and regained tremendous research interests recently due to the emergence of commodity depth cameras~\cite{supancic2015depth}. The problem is challenging due to the highly articulated structure, significant self-occlusion and viewpoint changes.

Existing methods can be categorized as two complementary paradigms, \emph{model based (generative)} or \emph{learning based (discriminative)}. Model based methods synthesize the image observation from hand geometry, define an energy function to quantify the discrepancy between the synthesized and observed images, and optimize the function to obtain the hand pose~\cite{oikonomidis2011efficient,qian2014realtime,makris2015hierarchical,tagliasacchi2015robust}. The obtained pose could be highly accurate, at the expense of dedicated optimization~\cite{sharp2015accurate}.

Learning based methods learn a direct regression function that maps the image appearance to hand pose, using either random forests~\cite{keskin2012hand,tang2013real,xu2013efficient,sun2015cascaded,li20153d} or deep convolutional neutral networks~\cite{tompson2014real,oberweger2015hands,oberweger2015training}. Evaluating the regression function is usually much more efficient than model based optimization. The estimated pose is coarse and can serve as an initialization for model based optimization~\cite{tompson2014real,poier2015hybrid,sridhar2015fast}.

Most learning based methods do not exploit hand geometry such as kinematics and physical constraints. They simply represent the hand pose as a number of independent joints. Thus, the estimated hand joints could be physically invalid, e.g., the joint rotation angles are out of valid range and the phalange length varies during tracking the same hand. Some works alleviate this problem via a post processing, e.g., using inverse kinematics to optimize a hand skeleton from the joints~\cite{tompson2014real,dong2015american}. Such post-processing is separated from training and is sub-optimal.

Recently, the deep-prior approach~\cite{oberweger2015hands} exploits PCA based hand pose prior in deep convolutional network. It inserts a linear layer in the network that projects the high dimensional hand joints into a low dimensional space. The layer is initialized with PCA and trained in the network in an end-to-end manner. The approach works better than its counterpart baseline without using such prior. Yet, the linear projection is only an approximation because the hand model kinematics is highly non-linear. It still suffers from invalid hand pose problem.

In this work, we propose a model based deep learning approach that fully exploits the hand model geometry. We develop a new layer that realizes the non-linear forward kinematics, that is, mapping from the joint angles to joint locations. The layer is efficient, differentiable, parameter-free (unlike PCA) and servers as an intermediate representation in the network. The network is trained end-to-end via standard back-propagation, in a similar manner as in~\cite{oberweger2015hands}, using a loss function of joint locations.

\begin{figure*}
\begin{center}
\includegraphics[width=0.9\linewidth]{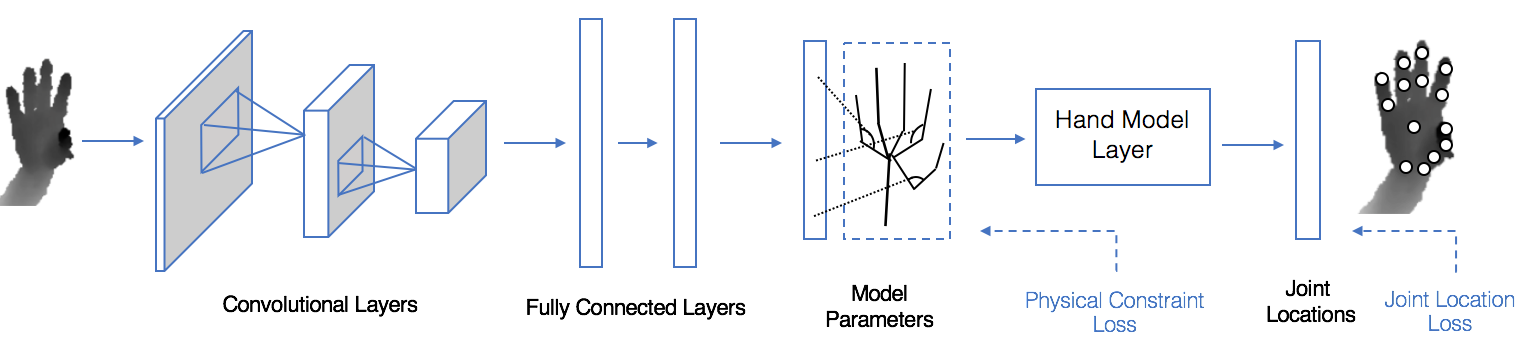}
\end{center}
   \caption{Illustration of model based deep hand pose learning. After standard convolutional layers and fully connected layers, the hand model pose parameters (mostly joint angles) are produced. A new hand model layer maps the pose parameters to the hand joint locations via a forward kinematic process. The joint location loss and a physical constraint based loss guide the end-to-end learning of the network.}
\label{fig:framework}
\end{figure*}

Our contributions are as follows:

\begin{itemize}
\item For the first time, we show that the end-to-end learning using the non-linear forward kinematics layer in a deep neutral network is feasible. The prior knowledge in the generative model of hand geometry is fully exploited. The learning is simple, efficient and gets rid of the inconvenient and sub-optimal post processing as in previous methods. The estimated pose is geometrically valid and ready for use.
\item Our approach is validated on challenging public datasets. It achieves state-of-the-art accuracy on both joint location and rotation angles. Specifically, we show that using joint location loss and adding an additional regularization loss on the intermediate pose representation are important for accuracy and pose validity.
\end{itemize}

The framework of our approach is briefly illustrated in Figure~\ref{fig:framework}. Our code is public available at {\tt https://github.com/tenstep/DeepModel}

\section{Related Work}

A good review of earlier hand pose estimation work is in ~\cite{erol2007vision}. ~\cite{supancic2015depth} provides an extensive analysis of recent depth based methods and datasets. Here we focus on the hybrid discriminative and generative approaches that are more related to our work. We also discuss other approaches that formulate handcraft operations into differentiable components.

\textbf{Hybrid approaches on hand pose} Many works use discriminative methods for initialization and generative methods for refinement. ~\cite{tompson2014real} predicts joint locations with a convolutional neural network. The joints are converted to a hand skeleton using an Inverse Kinematics(IK) process. ~\cite{sridhar2015fast} uses a pixel classification random forest to provide a coarse prediction of joints. Thus a more detailed similarity function can be applied to the following model fitting step by directly comparing the generated joint locations to the predicted joint locations. Similarly, ~\cite{poier2015hybrid} firstly uses a random regression forest to estimate the joint distribution, and then builds a more reliable quality measurement scheme based on the consistency between generated joint locations and the predicted distribution. All these approaches separate the joint estimation and model fitting in two stages. Recently, ~\cite{oberweger2015training} trains a feedback loop for hand pose estimation using three neutral networks. It combines a generative network, a discriminative pose estimation network and a pose update network. The training is complex. Our method differs from above methods in that it uses a single network and seamlessly integrates the model generation process with a new layer. The training is simple and results are good.

\textbf{Non-linear differentiable operations} In principle, a network can adopt any differentiable functions and be optimized end-to-end using gradient-descent. ~\cite{loper2014opendr} proposed a differentiable render to generate RGB image given appearance, geometry and camera parameters. This generative process can be used in neutral network. ~\cite{chiu2015see} leverages the fact that associated feature computation is piecewise differentiable, therefore Histogram of Oriented Gradient (HOG) feature can be extracted in a differentiable way. ~\cite{kontschieder2015deep} reformulates the split function in decision trees as a Bernoulli routing probability. The decision trees are plugged at the end of a neural network and trained together. As we know, we are the first to adopt a generative hand model in deep learning.

\section{Model Based Deep Hand Pose Estimation}

\subsection{Hand Model}

\begin{figure}
\begin{center}
\includegraphics[width=0.6\linewidth]{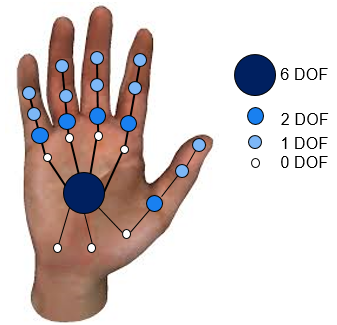}
\end{center}
\caption{Illustration of our hand model. It is similar to [Tompson \emph{et al}., 2014]. The hand pose is $26$ degrees of freedom (DOF), defined on $23$ internal joints. }
\label{fig:model}
\end{figure}

Our hand model is from libhand~\cite{libhand}. As illustrated in Figure~\ref{fig:model}, the hand pose parameters $\Theta\in\mathcal{R}^D$ have $D=26$ degrees of freedom (DOF), defined on 23 joints. There are $3$ DOF for global palm position, $3$ DOF for global palm orientation. The remaining DOF are rotation angles on joints.

Without loss of generality, let the canonical pose in Figure~\ref{fig:model} be a zero vector, the pose parameters are defined as relative to the canonical pose. Each rotation angle $\theta_i\in\Theta$ has a range $[\underline{\theta_i}, \overline{\theta_i}]$, which are the lower/upper bounds for the angle. Such bounds avoid self-collision and physically infeasible poses. They can be set according to the anatomical studies~\cite{albrecht2003construction}. In our experiments, they are estimated from the ground annotation on training data and provided in our published code.

We assume the bone lengths are known and fixed. Learning such parameters in a neutral network could be problematic as the results on the same hand could vary during tracking. Ideally, such parameters should be optimized once and fixed for each hand in a personal calibration process~\cite{khamis2015learning}. In our experiment, the bone lengths are set according to the ground truth joint annotation in NYU training dataset~\cite{tompson2014real}.

From $\Theta$ and bone lengths, let the forward kinematic function $\mathcal{F}: \mathcal{R}^D \rightarrow \mathcal{R}^{J \times 3}$ map the pose parameters to $J$ 3D joints ($J = 23$ in Figure~\ref{fig:model}). The kinematic function is defined on the hand skeleton tree in Figure~\ref{fig:model}. Each joint is associated with a local 3D transformation (rotation from its rotation angles and translation from its out-coming bone length). The global coordinate of a joint is obtained by transforming the origin via a series of the local transformations along the path from the hand root joint to the joint under consideration. The implementation details are provided in Appendix.

The forward kinetic function $\mathcal{F}$ is differentiable and can be used in a neutral network for gradient-descent like optimization. Yet, it is highly non-linear and its behavior during optimization could be different from the other linear layers in the network. In this work, we show that it is feasible to use such a non-linear layer during deep neutral network training.

\subsection{Deep Learning with a Hand Model Layer}
Taking an input depth image, our approach outputs the 3D hand joints and hand pose parameters $\Theta$. We use the same pre-processing as in previous work~\cite{oberweger2015hands,oberweger2015training}, assuming the hand is already detected (this can be done by a pixel-level classification random forest~\cite{tompson2014real} or assuming the hand is the closet object to the camera~\cite{qian2014realtime}). A fixed-size cube around the hand is extracted from the raw depth image. The spatial size is resized to  $128 \times 128$ and the depth values are normalized to $[-1,1]$.

Our network architecture is similar to the baseline network in deep prior approach~\cite{oberweger2015hands}, mostly for the purpose of fair comparison. It is illustrated in Figure~\ref{fig:framework}. It starts with 3 convolutional layers with kernel size $5, 5, 3$, respectively, followed by max pooling with stride $4, 2, 1$ (no padding), respectively. All the convolutional layers have 8 channels. The result convolutional feature maps are $12 \times 12 \times 8$. There are then two fully connected (fc) layers, each with 1024 neurons and followed by a dropout layer with dropout ratio 0.3. For all convolutional and fc layers, the activation function is ReLU.

After the second fc layer, the third fc layer outputs the $26$ dimensional pose parameter $\Theta$. It is connected to a hand model layer that uses the forward kinematic function $\mathcal{F}$ to output the 3D joint locations. A Euclidian distance loss for the joint location is at last. Unlike~\cite{tompson2014real,oberweger2015hands}, we do not directly output the joint locations from the last fc layer, but use an intermediate hand model layer instead, which takes hand geometry into account and ensures the geometric validity of output.

The joint location loss is standard Euclidian loss.
\begin{equation} \label{jt}
L_{jt}(\Theta) = \frac{1}{2} ||\mathcal{F}(\Theta) - Y||^2
\end{equation}
, where $Y \in \mathcal{R}^{J \times 3}$ is the ground truth joint location.

We also add a loss that enforces the physical constraint on the rotation angle range, as
\begin{equation} \label{phy}
L_{phy}(\Theta) = \sum_i{[max(\underline{\theta_i} - \theta_i, 0) + max(\theta_i - \overline{\theta_i}, 0)]}.
\end{equation}

Therefore, the overall loss with respect to the pose parameter $\Theta$ is
\begin{equation} \label{overall}
L(\Theta) = L_{jt}(\Theta) + \lambda L_{phy}(\Theta)
\end{equation}
, where weight $\lambda$ balances the two loss and is fixed to 1 in all our experiments.

In optimization, we use standard stochastic gradient descent, with batch size 512, learning rate 0.003 and momentum 0.9. The training is processed until convergence.

\subsection{Discussions}

In principle, any differentiable functions can be used in the network and optimized via gradient descent. Yet, for non-linear functions it is unclear how well the optimization can be done using previous practices, such as parameter setting. Our past experiences in network training are mostly obtained from using non-linearities like ReLu or Sigmoid. They are not readily applicable for other non-linear functions.

Our experiment shows that our proposed network is trained well. We conjecture a few reasons. Our hand model layer is parameter free and has no risk of over-fitting. The gradient magnitude of the non-linear 3D transformation (mostly $\sin$ and $\cos$) is well behaved and in stable range (from $-1$ to $1$). The hand model layer is at the end of the network and does not interfere with the previous layers too much. Our approach can be considered as transforming the last Euclidian loss layer into a more complex loss layer when combining the last two layers together.

The joint loss in (\ref{jt}) is well behaved as the errors spread over different parts. This is important for learning an articulated structure like hand. Intuitively, roles of different dimensions in pose parameter $\Theta$ are quite different. The image observation as well as joint locations are more sensitive to the global palm parameters (rotation and position) than to the finger parameters. This makes direct estimation of $\Theta$ hard to interpret and difficult to tune. In experiment, we show that using joint loss is better than directly estimating $\Theta$.

The physical constraint loss in (\ref{phy}) helps avoiding invalid poses, as verified in the experiment.

\section{Experiment Evaluation}

\begin{figure}[!h]
\begin{center}
\includegraphics[width=0.8\linewidth]{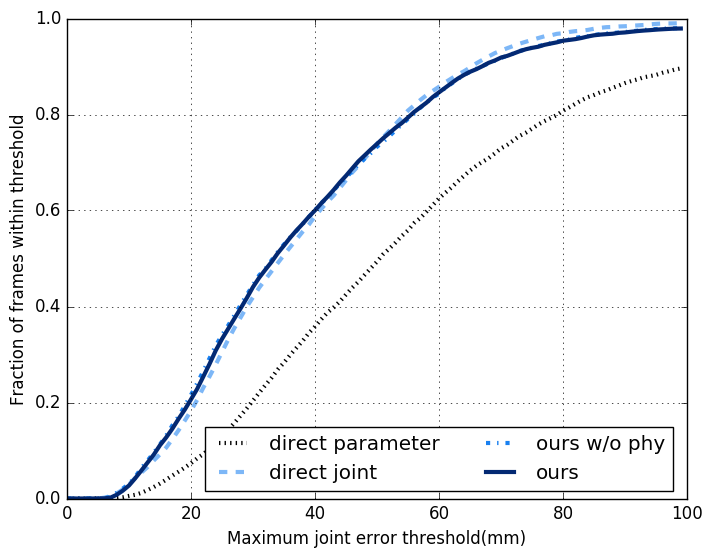}
\includegraphics[width=0.8\linewidth]{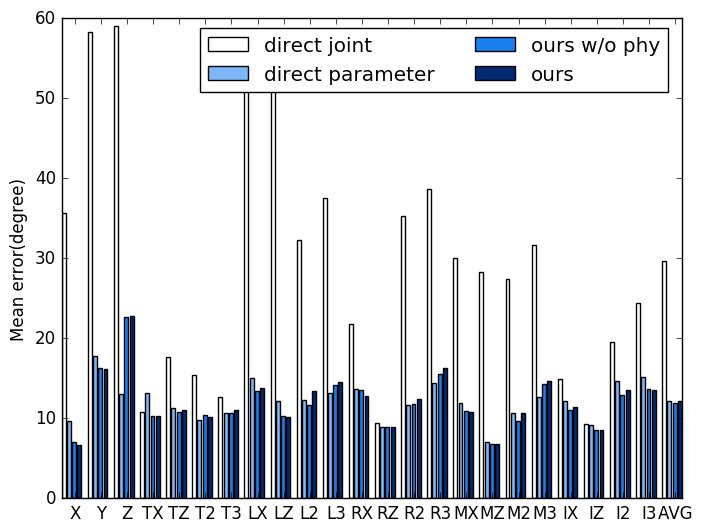}
\end{center}
   \caption{Comparison of our approach and different baselines on NYU test dataset. The upper shows the fraction of frames with maximum joint error below certain thresholds. The lower shows the average error on individual angles.}
\label{fig:selfNYU}
\end{figure}

Our approach is implemented in Caffe~\cite{jia2014caffe}. The hand model layer is efficient enough and performed on the CPU. On a PC with an Intel Core i7 4770 3.40GHZ, 32GB of RAM, and an Nvidia GeForce 960 GPU, one forward pass takes about 8ms, resulting in 125 frames per second in test.

We use two recent public datasets that are widely used in depth based hand pose estimation.

{\bf NYU}~\cite{tompson2014real} dataset contains 72757 training and 8252 testing images, captured by PrimeSense camera. Ground truth joints are annotated using an accurate offline particle swarm optimization (PSO) algorithm, similar to ~\cite{oikonomidis2011efficient}. As discussed in ~\cite{supancic2015depth}, NYU dataset has the largest pose variation and is the most challenging in all public hand pose datasets. It is therefore used for our main evaluation.

NYU dataset's ground truth 3D joint location is accurate. Although 36 joints are annotated, evaluation is only performed on a subset of 14 joints, following previous work~\cite{tompson2014real,oberweger2015hands}. For more rigorous evaluation, we also obtain the ground truth hand pose parameters from ground truth joints. Similarly as in~\cite{tang2015opening}, we apply PSO to find the ground truth pose $\Theta$ (frame by frame) that minimizes the loss in Equation ~\eqref{jt} with $J = 14$. To verify the accuracy of such estimated poses, we compare the original ground truth joints with the joints computed from our optimized poses (via forward kinematic function $\mathcal{F}$). The average error is $5.68 mm$ and variance is $1.94 mm^2$, indicating an accurate fitting using our hand model.

{\bf ICVL}~\cite{tang2014latent} dataset has over 300k training depth images and 2 testing sequences with each about 800 frames. The depth images are captured by Intel Creative Interactive Gesture Camera. However, its ground truth joint annotation is quite inaccurate. We use this dataset mainly for completeness as some previous works use it.

We use three evaluation metrics. The first two are on joint accuracy and used in previous work~\cite{oberweger2015hands,oberweger2015training,tang2014latent}. First is the average joint error over all test frames. Second, as a more challenging and strict metric, is the proportion of frames whose maximum joint error is below a threshold.

In order to evaluate the accuracy of pose estimation, we use the average joint rotation angle error (over all angles in $\Theta$) as our last metric.

\begin{table}
\begin{center}
\begin{tabular}{|l|*{3}{c|}}\hline
\backslashbox{Methods}{Metrics} & Joint location error & Angle error\\\hline
{direct joint} & $17.2mm$ & $21.4^{\circ}$ \\\hline
{direct parameter} & $26.7mm$ & $12.2^{\circ}$ \\\hline
{ours w/o phy} & $\bf 16.9mm$ & $\bf 12.0^{\circ}$ \\\hline
{ours} & $\bf 16.9mm$ & $12.2^{\circ}$ \\\hline
\end{tabular}
\caption{Comparison of our approach and different baselines on NYU test dataset. It shows that our approach is best on both average joint and pose (angle) accuracy.}
\label{tab:selfNYU}
\end{center}
\end{table}

\subsection{Evaluation of Our Approach}
Our approach uses an intermediate model based layer. Learning is driven by joint location loss. To validate its effectiveness, it is compared to two baselines. The first one directly estimates the individual joints. It is equivalent to removing the model parameters and hand model layer in Figure~\ref{fig:framework}. It is actually the baseline in deep prior approach~\cite{oberweger2015hands}. We refer this baseline as \textbf{direct joint}. The second one is similar to first one, except that the regression target is not joint location but the pose parameters (the global position and rotation angles in $\Theta$)\footnote{We also experimented with adding the physical constraint loss but observed little difference.}. We refer this baseline as \textbf{direct parameter}. Note that this baseline is trained using the ground truth pose parameters we obtained, as described earlier. Further, we refer our approach without using the physical constraint loss in Equation \eqref{phy} as \textbf{ours w/o phy}.

\begin{figure*}
\begin{center}
\includegraphics[width=0.9\linewidth]{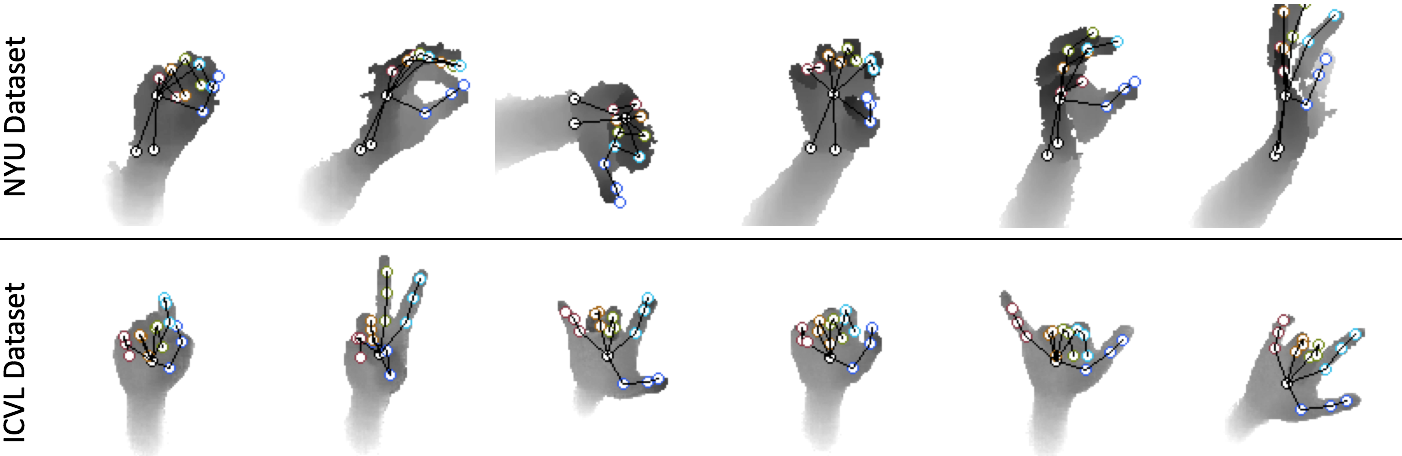}
\end{center}
\caption{Example results on NYU and ICVL datasets. The estimated 3D joints are overlaid on the depth image. Our method is robust to various viewpoints and self-occlusion.}
\label{fig:qualitative}
\end{figure*}

As shown in Figure~\ref{fig:selfNYU} and Table~\ref{tab:selfNYU}, our approach is the best in terms of all evaluation metrics, demonstrating that the hand model layer is important to achieve good performance for both joint and pose parameter estimation.

For the direct joint approach, we estimate the angle parameters using the similar PSO based method described above. That is, the pose parameters are optimized to fit the estimated joints, in a post-processing step. As the direct joint learning does not consider geometric constraints, one can expect that such fitting for the model parameters is poor. Indeed, the average difference between the optimized joint angles and ground truth joint angles is large, indicating that the estimated joints in many frames are geometrically invalid, although the joint location errors are relatively low (see Table~\ref{tab:selfNYU}).

Direct parameter approach has decent accuracy on angles since that is the learning objective. However, it has largest joint location error, probably because small error in angle parameters does not necessarily imply small error in joint location. For example, a small error in global rotation could result in large error in finger tips, even when the finger rotation angles are accurate.
\begin{figure}
\begin{center}
\includegraphics[width=0.8\linewidth]{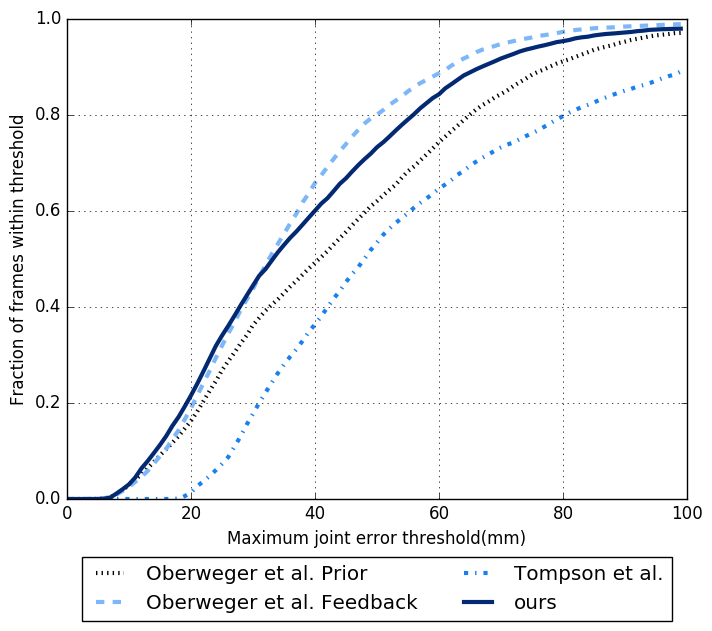}
\end{center}
\caption{Comparison of our approach and state-of-the-art methods on NYU test dataset. It shows the fraction of frames with maximum joint error below certain thresholds.}
\label{fig:others_within_NYU}
\end{figure}
In ours w/o phy, we have best performance on both joints and rotation angles. Yet, when we consider the joint angle constraint, we find that in $18.6\%$ of the frames, there is at least one estimated angle out of the valid range. When using physical constraint loss (ours), this number is reduced to $0.9\%$, and accuracy on both joints and rotation are similar (Table~\ref{tab:selfNYU}). These results indicate that 1) using a hand model layer with a joint loss is effective; 2) the physical constraint loss ensures the geometric validity of the pose estimation.

Example results of our approach are shown in Figure~\ref{fig:qualitative}.

\subsection{Comparison with the State-of-the-art}
\begin{figure}
\begin{center}
\includegraphics[width=0.8\linewidth]{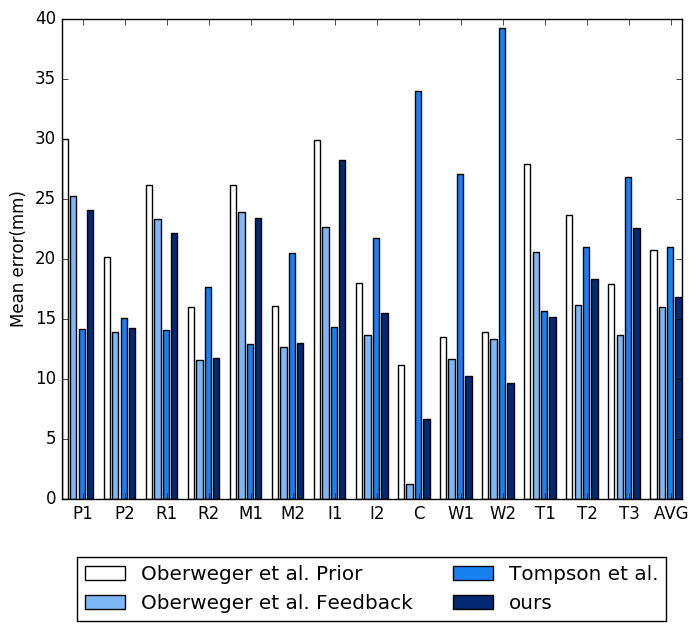}
\end{center}
\caption{Comparison of our approach and state-of-the-art methods on NYU test dataset.It shows the average error on individual joints.}
\label{fig:others_mean_NYU}
\end{figure}

In this section, we compare our method with the state-of-the-art methods. For these methods, we use their published original result.

On the NYU dataset, our main competitors are ~\cite{tompson2014real,oberweger2015hands}. Both are based on convolutional neural networks and are similar to our direct joint baseline. We also compare with~\cite{oberweger2015training}. It trains a feedback loop that consists of three convolutional neural networks. It is more complex and is currently the best method on NYU dataset. Results in Figure~\ref{fig:others_within_NYU} and Figure~\ref{fig:others_mean_NYU} show that our approach clearly outperforms~\cite{tompson2014real,oberweger2015hands} and is comparable with~\cite{oberweger2015training}.

\begin{figure} [!h]
\begin{center}
\includegraphics[width=0.8\linewidth]{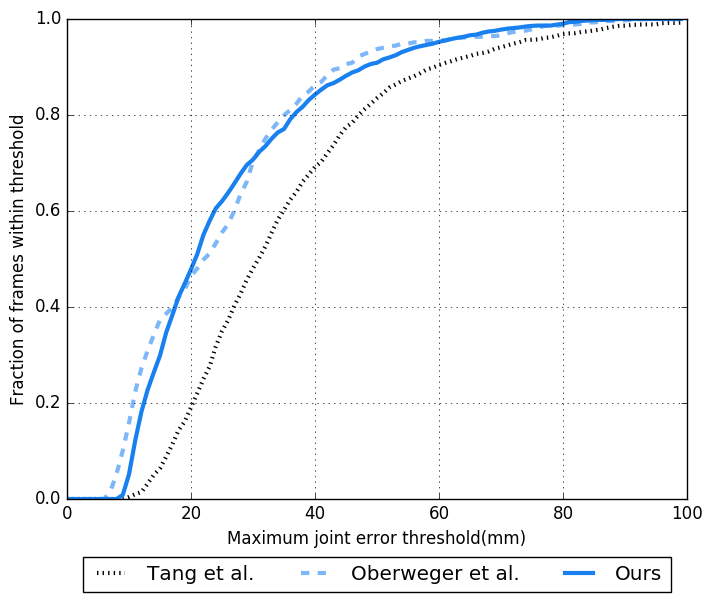}
\includegraphics[width=0.8\linewidth]{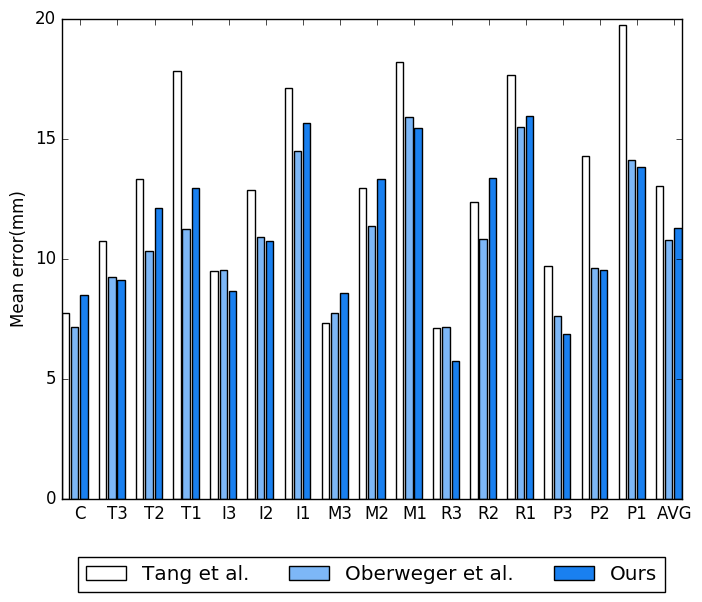}
\end{center}
   \caption{Comparison of our approach and state-of-the-art methods on ICVL test dataset. It shows the fraction of frames with maximum joint error below certain thresholds and the average error on individual joints.}
\label{fig:others_within_ICVL}
\end{figure}

On the ICVL dataset, we compare with ~\cite{tang2014latent} and ~\cite{oberweger2015hands}. Results in Figure~\ref{fig:others_within_ICVL} show that our method significantly outperforms ~\cite{tang2014latent} and is comparable with~\cite{oberweger2015hands}. We note that the ICVL dataset has quite inaccurate joint annotation and small viewpoint changes (as discussed in~\cite{supancic2015depth}). Both are disadvantageous for our model based approach because it is more difficult to fit a model to inaccurate joints and the strong geometric constraints enforced by the model are less effective in near-frontal viewpoints. We also note that we use the same geometric hand model as for NYU dataset and only learn the rotation angles. Considering such limitations, our result on ICVL is quite competitive.

\section{Conclusions}

We show that it is possible to integrate the forward kinematic process of an articulated hand model into the deep learning framework for effective hand pose estimation. Such an end-to-end training is clean, efficient and gets rid of the inconvenient post-processing used in previous approach. Extensive experiment results verify the state-of-the-art performance of proposed approach.

Essentially, our approach exploits the prior knowledge in geometric hand model in the learning process. It can be easily applied to any articulated pose estimation problem such as human body. More broadly speaking, any deterministic and differentiable generative model can be used in a similar manner~\cite{loper2014opendr,oberweger2015training}. We hope this work can inspire more works on effective integration of generative and discriminative methods.

\section*{Acknowledgments}
This work was supported in part by NSFC under Grant 61473091 \& 61572138, two STCSM’s programs (No. 15511104402 \& No. 15JC1400103), and EU FP7 QUICK Project under Grant Agreement (No.PIRSESGA2013-612652).

\section*{Appendix on hand model kinematics}

\begin{figure}
\begin{center}
\includegraphics[width=0.75\linewidth]{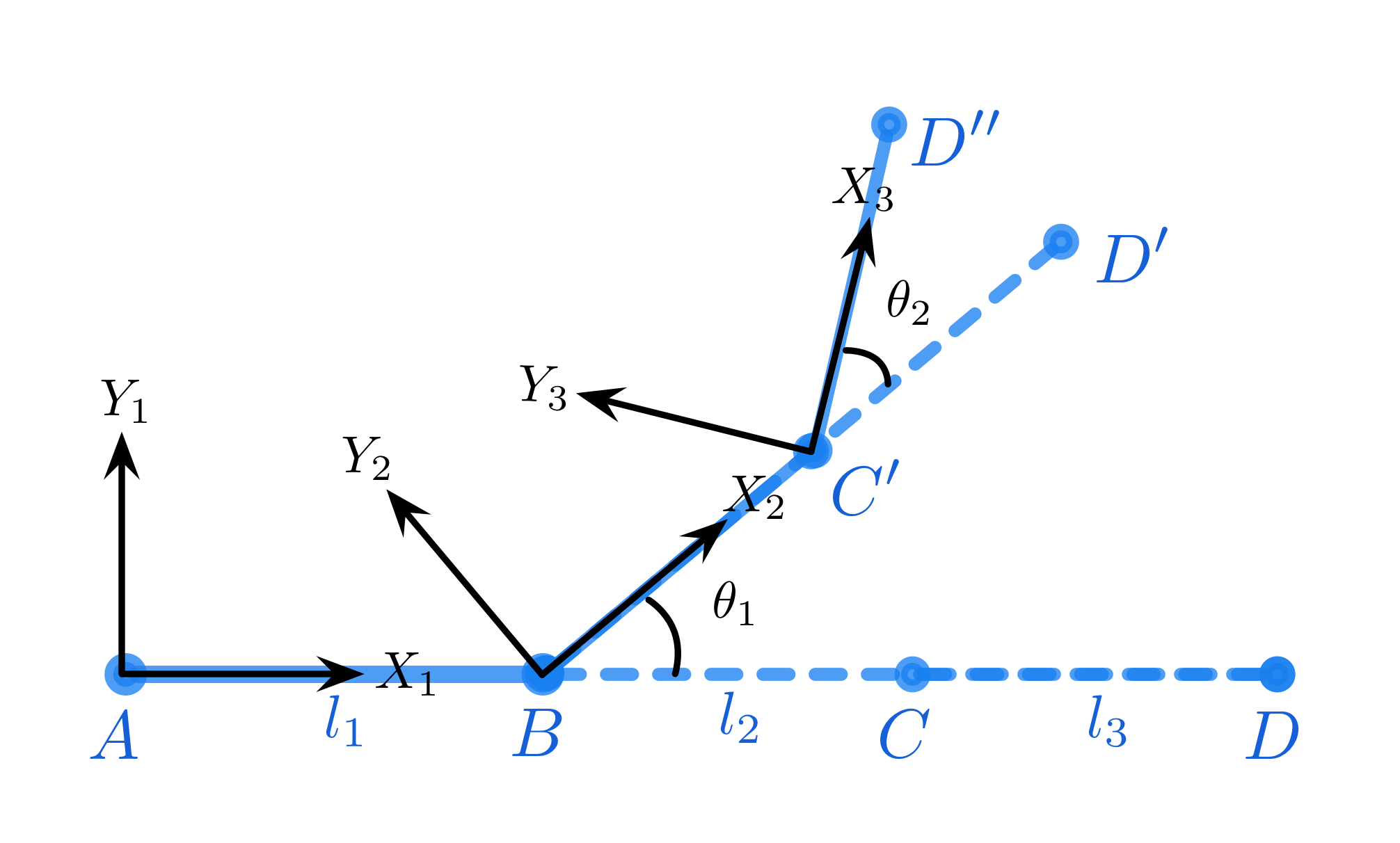}
\end{center}
   \caption{Illustration of forward kinematic implementation. Joint $A, B, C, D$ are 4 adjacent joints of the initial hand model(not necessarily collinear). The relative 3D coordinate of joint $D''$ with respect to A after two rotations centered at joint B and C among axis Z can be written as $\textbf{p}_{D''} = Trans_x(l_1) \times Rot_z(\theta_1) \times Trans_x(l_2) \times Rot_z(\theta_2) \times Trans_x(l_3) \times [0, 0, 0, 1]^\top$}
\label{fig:rotation}
\end{figure}

The hand model layer takes the model parameters as input and outputs the corresponding joint coordinates. As the hand model is a tree struct kinematic chain, the transformation of each joint is a forward-kinematic process. We can consider the transformation of two adjacent joints as transform two local coordinate systems. Let the original coordinate of a point be $(0, 0, 0)$, represented in homogenous coordinate $[0, 0, 0, 1]^\top$, $Trans_{\phi}(l)$ is the 4x4 transformation matrix that transforms $l$ among axis $\phi \in \{X, Y, Z\}$, and $Rot_{\phi}(\theta)$ is the 4x4 rotation matrix that rotate $\theta$ degrees among axis $\phi$. Generally, let $Pa(u)$ be the set of parent joints of joint $u$ on the kinematic tree(rooted at hand center), the coordinate of $u$ after $k$ relative rotation is:
\begin{equation} \label{chain}
\textbf{p}_{u^{(k)}} = (\prod_{t \in Pa(u)} Rot_{\phi_t}(\theta_t) \times Trans_{\phi_t}(\theta_t)) [0, 0, 0, 1] ^ \top
\end{equation}
Note that most of the joints have more than one rotation DOF, but the formulation is the same as equation \eqref{chain}, as the additional rotation matrices are multiplied on the left of the corresponding joints.
The derivation of joint coordinate u with respect to joint angle t is replace the rotation matrix of joint angle t(if exists) by it's derivation and keep other matrix unchanged.

\small
\bibliographystyle{named}
\bibliography{ijcai16}

\end{document}